\documentclass{article}

\usepackage{arxiv}

\usepackage[utf8]{inputenc} 
\usepackage[T1]{fontenc}    
\usepackage{hyperref}       
\usepackage{url}            
\usepackage{booktabs}       
\usepackage{amsfonts}       
\usepackage{nicefrac}       
\usepackage{microtype}      
\usepackage{lipsum}		
\usepackage{graphicx}
\usepackage{natbib}
\usepackage{doi}
\usepackage{textcomp}
\usepackage{graphicx}
\usepackage{amsmath}
\usepackage[version=4]{mhchem}
\usepackage{siunitx}
\usepackage{threeparttable}
\usepackage{longtable,tabularx}
\usepackage{diagbox}
\usepackage{tabularx}

\title{Hospital transfer risk prediction for COVID-19 patients from a medicalized hotel based on Diffusion GraphSAGE}

\author{ \href{https://orcid.org/0000-0002-1233-138X}{\includegraphics[scale=0.06]{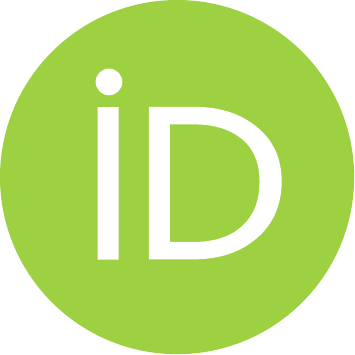}\hspace{1mm}Jun-En Ding}\thanks{Use footnote for providing further
		information about author (webpage, alternative
		address)---\emph{not} for acknowledging funding agencies.} \\
	Institute of Hospital and Health Care Administration\\
        National Yang Ming Chiao Tung University \\Taipei City, Taiwan  \\
	\texttt{m02040013@nycu.edu.tw} \\
        \And
	\href{https://orcid.org/0000-0003-3359-5429}{\includegraphics[scale=0.06]{orcid.pdf}\hspace{1mm}Chih-Ho Hsu} \\
	Far Eastern Memorial Hospital \\
         Banqiao District, New Taipei City, Taiwan\\
        \texttt{chihhohsu.femh@gmail.com}\\
        \And
         \href{https://orcid.org/0000-0001-6877-5345}{\includegraphics[scale=0.06]{orcid.pdf}\hspace{1mm}Kuan-Yin Lin} \\
          Institute of Hospital and Health Care Administration \\
          National Yang Ming Chiao Tung University \\Taipei City, Taiwan \\
          \texttt{kuanyin0828@gmail.com} \\
          \And
       \href{https://orcid.org/0000-0002-1512-6672}{\includegraphics[scale=0.06]{orcid.pdf}\hspace{1mm}Ling Chen} \\
        Institute of Hospital and Health Care Administration \\
        National Yang Ming Chiao Tung University \\Taipei City, Taiwan  \\
	\texttt{ling.chen@nycu.edu.tw}
        \And
	\href{https://orcid.org/0000-0003-3501-5459}{\includegraphics[scale=0.06]{orcid.pdf}\hspace{1mm}Fang-Ming Hung} \\
	Far Eastern Memorial Hospital \\
	Banqiao District, New Taipei City, Taiwan \\
	\texttt{philip@mail.femh.org.tw}}

\renewcommand{\shorttitle}{\textit{arXiv} Template}

\hypersetup{
pdftitle={A template for the arxiv style},
pdfsubject={q-bio.NC, q-bio.QM},
pdfauthor={David S.~Hippocampus, Elias D.~Striatum},
pdfkeywords={First keyword, Second keyword, More},
}

\begin{document}
\maketitle

\begin{abstract}
The global COVID-19 pandemic has caused more than six million deaths worldwide. Medicalized hotels were established in Taiwan as quarantine facilities for COVID-19 patients with no or mild symptoms. Due to limited medical care available at these hotels, it is of paramount importance to identify patients at risk of clinical deterioration. This study aimed to develop and evaluate a graph-based deep learning approach for progressive hospital transfer risk prediction in a medicalized hotel setting. Vital sign measurements were obtained for 632 patients and daily patient similarity graphs were constructed. Inductive graph convolutional network models were trained on top of the temporally integrated graphs to predict hospital transfer risk. The proposed models achieved AUC scores above 0.83 for hospital transfer risk prediction based on the measurements of past 1, 2, and 3 days, outperforming baseline machine learning methods. A post-hoc analysis on the constructed diffusion-based graph using Local Clustering Coefficient discovered a high-risk cluster with significantly older mean age, higher body temperature, lower SpO\textsubscript{2}, and shorter length of stay. Further time-to-hospital-transfer survival analysis also revealed a significant decrease in survival probability in the discovered high-risk cluster. The obtained results demonstrated promising predictability and interpretability of the proposed graph-based approach. This technique may help preemptively detect high-risk patients at community-based medical facilities similar to a medicalized hotel.
\end{abstract}


\section{Introduction}
The novel coronavirus disease 2019 (COVID-19) has become a pandemic worldwide, imposing a heavy burden on the healthcare systems. To manage the rapidly growing number of patients, temporary health care facilities, such as Fangcang shelter hospitals in China \cite{Fangcang}, were established in many countries \cite{10.1093/milmed/usab003, Bergamo, Bulajic, Pandemic_Care}. In Taiwan, medicalized hotels were set up as a type of quarantine facility after a Level 3 COVID-19 alert was announced on 19 May 2021, by the Central Epidemic Command Center (CECC). Due to the limited resources and facilities in medicalized hotels, it was crucial to identify potentially severe patients early and transfer them to a hospital in time for better care. Particularly, patients experiencing rapid clinical deterioration, defined as having oxygen saturation (SpO\textsubscript{2}) less than 94\% or respiratory rate more than 30 breaths per minute, according to the COVID-19 treatment guidelines of the National Institutes of Health \cite{Underlying_Medical}, were considered as at high risk.

Machine learning (ML) is a family of algorithms that learn from data for analytic or predictive tasks \cite{ISLR}. Existing COVID-19 related ML prediction models mainly focused on severity \cite{Annals_Haimovich,Wu2001104,MLCOVID}, mortality \cite{predCOVID-19,mortality_risk,Epidemiology}, and survival time \cite{NEMATI2020100074,DICASTELNUOVO20201899} predictions. In recent years, deep learning (DL), a subdomain of ML based on artificial neural networks \cite{Goodfellow}, has shown promising performance in a wide range of applications. Deep convolutional neural networks models were introduced to diagnose COVID patients \cite{Mohamadou2020}, mostly based on computed tomography (CT) scans \cite{Jangam2022,QI2021106406,Integrating}, X-rays \cite{KHAN2020105581,P2021}, and intensive care unit (ICU) data \cite{Incremental}.

Graph representation is commonly used to capture objects and their relationships as nodes and edges, for example, gene-phenotype \cite{Genome-wide,Wang2014SimilarityNF} and gene-disease \cite{Wang2014SimilarityNF} relationships. Graph Convolutional Networks (GCNs), originally evolved from Graph Neural Networks, have gained increasing attention in recent years \cite{KipfW16}. Several studies adapted GCNs for COVID-19 related tasks, such as COVID-19 diagnosis and severity prediction based on CT scans \cite{keicher2021ugat, YU2021592} or laboratory data \cite{Ferrari}. For example, Ferrari et al \cite{Ferrari} proposed a Bayesian Structure Learning-based framework that learned a probabilistic directed acyclic graph from clinical data, such as past history and blood analysis, to identify the main causes of patient outcome. Graph representation was also used to capture social network connectivity for COVID-19 transmission based on social network data \cite{KRI}. However, most existing studies relied on laboratory data or medical images \cite{pneumonia_diagnosis, covid_detection,Boundary}, which were not available in medicalized hotels.

\begin{figure}
    \centering
    \includegraphics[width=0.9\linewidth]{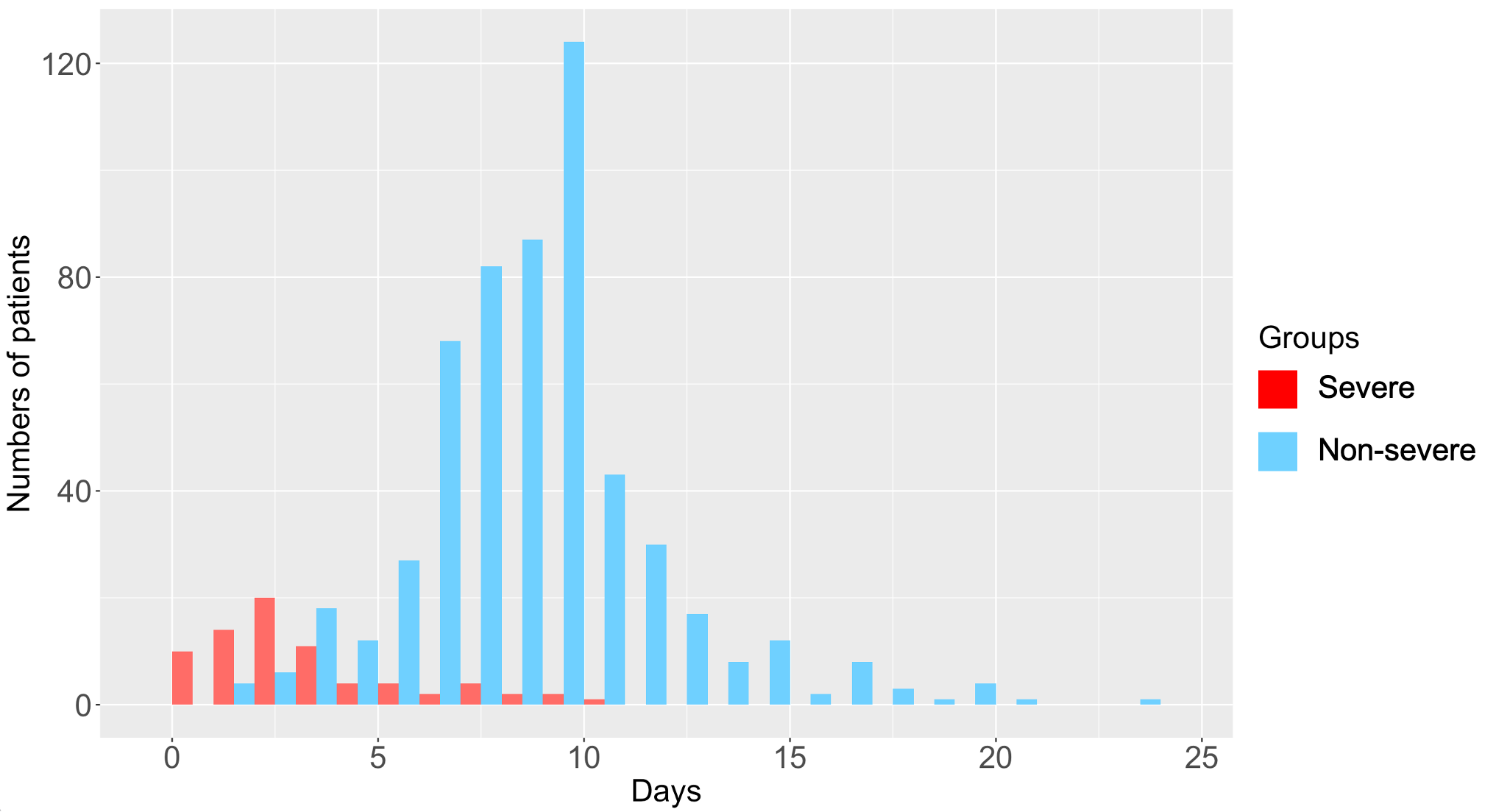}
    \caption{Daily patient count, with severe patients (being transferred on the day) colored in red and non-severe patients (remaining at the hotel on the day) in light blue. Most severe patients were transferred to a hospital within the first 3-4 days.}
    \label{fig:Fig1}
 \end{figure}

 The current study aimed to develop and evaluate models for daily hospital transfer risk prediction based on limited patient vital signs data available at a medicalized hotel, with ML and graph-based deep learning approaches.

\section{Materials and Methods}\label{sec2}
\subsection{Study population and data collection protocol}

In this retrospective cohort study, our dataset contained 679 patients admitted to the Banqiao medicalized hotel in New Taipei City, Taiwan, between 28 May and 7 July 2021. The hotel was managed by the Far Eastern Memorial Hospital (FEMH). During the pandemic of the alpha-variant, patients with COVID symptoms were advised to visit a hospital emergency department (ED). If tested COVID positive with stable clinical condition, the patients were transferred to a designated quarantine hotel. When arriving at the hotel, the patients were evaluated in the ambulance for triage, and 47 patients with unstable vital signs were immediately transferred to a hospital for more adequate treatment. 632 patients in total were admitted to the Banqiao medicalized hotel during the 41 days of operation. The hotel used a virtual cloud-based ward care system for medical staff to execute daily ward rounds, medical consultations, and vital signs measurements on a social media platform. Basic vital signs were measured at least once per day, eight hours apart, including body temperature (BT), pulse rate (PR), peripheral oxygen saturation (SpO\textsubscript{2}), systolic blood pressure (SBP), and diastolic blood pressure (DBP).
Throughout the quarantine period, 68 patients were transferred to a hospital due to clinical deterioration and the majority were transferred within the first 3-4 days (Fig. \ref{fig:Fig1}). On Day 9 of admission, afebrile and asymptomatic patients were checked for the viral load by Reverse transcription PCR (RT-PCR) from the nasal swab specimen. If the Ct (cycle threshold) value was greater than 27, the patients would be discharged the next day and asked to stay at home for another 7-day isolation. Otherwise, they would stay up to 14 days in the medicalized hotel and another 7-day isolation at home. Finally, 538 patients were discharged from the medicalized hotel. No medical staff was infected with COVID-19 during this 41-day operation \cite{LIM2021}. 

\begin{figure}[t!]
    \centering
    \includegraphics[width=0.9\linewidth]{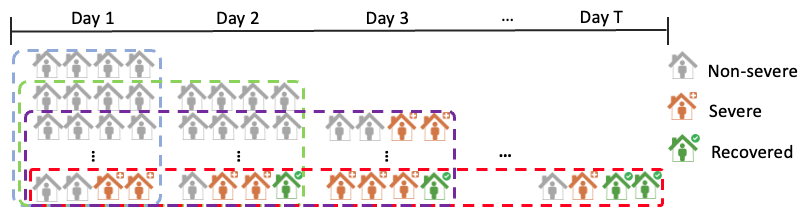}
    \caption{All patients were aligned at their arrival date. As the severe cases were transferred and non-severe cases were discharged along the time, the number of patients included in model development and testing decreased as the length of stay increased.}
    \label{fig:Fig2}
 \end{figure}

\subsection{Data preparation}
Patient data was aligned at the arrival day. We assigned a daily label to every patient based on whether they were transferred to a hospital on that day (i.e., a positive label if yes; otherwise, negative). Once a person left the medicalized hotel, either being discharged or transferred to a hospital, they were not included in the following days’ modelling or testing. This design is illustrated in Fig. \ref{fig:Fig2}, where the number of patients involved in the study decreases as the length of stay increases.

Missing values were observed in the daily vital sign measurements with missing rates 4.93\% (SpO\textsubscript{2}), 5.48\% (BT), 6.98\% (PR), 81.42\% (SDP), and 81.44\% (DBP). To handle missing values, we used Multivariate Imputation by Chained Equations (MICE algorithm) \cite{MICE}, which imputed the missing values of a feature at time $t$ based on the data before $t$.
Due to the data imbalance issue in our dataset, we used an oversampling technique called SMOTE \cite{dynamic} to increase the number of positive samples. 
The SMOTE algorithm selected a sample from the minority class and randomly selected $N$ samples from the $K$ nearest neighbors to synthesize $M$ new positive samples.

 \subsection{Model design}
\begin{figure}[t!]
    \centering
    \includegraphics[width=0.99\linewidth]{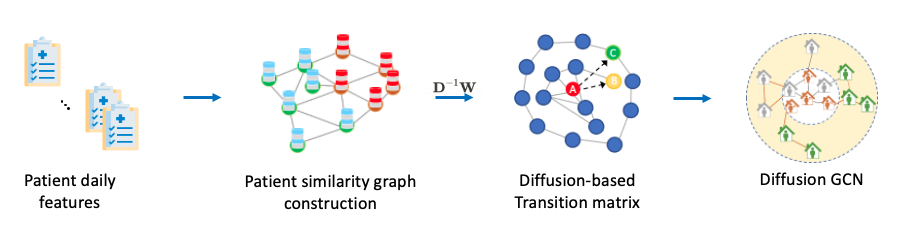}
    \caption{An overview of the proposed graph-based hospital transfer risk prediction model.}
    \label{fig:Fig3}
 \end{figure}

 In this study, we designed a graph convolutional networks approach for hospital transfer risk prediction. As patients were aligned at their arrival, a Day $T$ prediction refers to the use of the first $T$ days’ data to predict the hospital transfer risk on Day $T$. In other words, our goal was to build models capable of making predictions progressively based on the data collected so far. Unlike existing studies on COVID-19 related predictions, our models only used vital signs and explored the temporal dependencies for potential disease progression. An overview of the design can be found in Fig. \ref{fig:Fig3}. The main components of the proposed risk prediction framework included patient similarity graph construction and building an inductive GCN model.  
 
 \subsubsection{K-neighbors graph construction}
 We first constructed a patient similarity graph using the k-neighbors method. Let patient features on Day $t$ be represented as a matrix $\textbf{X}^{(t)}$ $\in$ $\mathbb{R}^{n \times p }$, $t=1,...,T$. Where $n$ is the number of patients included on Day $t$ and $p$ is the number of vital signs measurements. To capture similarities among patients on each day, we built a temporal weighted adjacency matrix using K-neighbors for the day. In a K-neighborhood, two data points $i$ and $j$ connected by an edge $(i,j)$ if $i$ is amongst the $K$ nearest neighbors of $j$ or vice versa. A weighted graph is defined as $\mathcal{G}$ $(V,E,\mathbf{W})$, where $V$ and $E$ are the sets of nodes and edges respectively, and $\mathbf{W}$ is a weighted matrix computed from a function of Gaussian kernel defined as below:

 \begin{center}
$
{W}_{ij}=W(X_{i},X_{j}) =
\begin{cases} 
exp(-\frac{\left\|X_{i} - X_{j} \right\|^2}{\alpha}),  & \mbox{if }X_{i} \in \mathcal{N}(X_{j})  \hspace{8pt} or \hspace{8pt}X_{j} \in \mathcal{N}(X_{i}) \\
0, & \mbox{if }X_{i} \not\in \mathcal{N}(X_{j}) \hspace{8pt} or \hspace{5pt} X_{j} \not\in \mathcal{N}(X_{i})
\end{cases}
$
\end{center}

\noindent where $W_{ij}$ is an undirected edge of vertices $i$ and $j$, $X_i$ denotes row $i$ in $\textbf{X}$, and $\alpha$ is a scale parameter.

\subsubsection{Diffusion graph construction}
We further extended the k-neighbors-based graph to a diffusion-based graph to capture disease progression over time. A transition matrix can be defined as $\textbf{D}^{-1}\textbf{W}$ where the degree matrix is defined as $\mathbf{D} =$
$\textbf{diag}{(d_{11},d_{22},...,d_{nn})}$, where $d_{ii}$ is the sum of row $i$ in  $W_{ij}$. To build a diffusion transition graph, given the transition matrices of the weighted graphs $\left[ \mathcal{G}^{(1)},...,  \mathcal{G}^{(t)}, ..., \mathcal{G}^{(T)} \right]$ defined previously for Day 1, ..., Day $T$, the diffusion distances \cite{LIM2021} on the graph can be computed as the linear combination of weighted graph of infinite random walk and stationary diffusion distribution $\mathcal{P}$ \cite{traffic}:

\[\mathcal{P}=\sum_{t=1}^{\infty}(\textbf{D}^{-1}\textbf{W})^{t} \]

Therefore, we can use the patient features up to Day $T$ as a $T$-step truncation process in a finite diffusion step. Alternatively, the $\mathcal{P}$ can be interpreted as the cumulative probability of the patient's state in a diffusion process on Day $T$. 

\subsubsection{Inductive GCNs}
Graph convolutional networks (GCNs) \cite{KipfW16} is a deep learning architecture commonly applied to graph node classification problems. Traditional graph convolutional networks belong to a transductive learning approach, where both training and testing data are used to construct the learning graph at the training time. However, a transductive approach is not practical in most medical settings, since new patient data is often unavailable at the model development time. To tackle this problem, we adapted a graph convolutional network called GraphSAGE \cite{NIPS2017_5dd9db5e} capable of inductively predicting the unseen nodes. Using a sampling strategy, the model was able to learn node embeddings based on neighboring node features rather than the entire graph. When an inference was performed, the learned information can be aggregated to the weight of the target node. Given a patient similarity graph constructed previously with the set of nodes $V = 	\left\{v_{1},..,v_{n}\right\}$, the model computed for every node its connectivity with its depth $k$ for $k=1,...,K$, using $K$ network layers. Let $\mathcal{N}(v)$ denote the neighborhood of $v$ and $h^{k-1}_{\mathcal{N}(v)}$ be the aggregated representation of $v$'s immediate neighborhood $\left\{h^{k-1}_u, \forall u \in {\mathcal{N}(v)}\right\}$ at depth $k-1$. Then the convolutional propagation at layer $k$ for node $v$ is defined based on the concatenation of its previous layer output and its current aggregated neighborhood: 

     \[h^{k}_{v} = \sigma(W^{k}\cdot CONCAT(h^{k-1}_{v}, h^{k}_{\mathcal{N}(v)})) \]
     
Let $z_v$ be the output of the final aggregated information from $h^{k}_{v}$, \; $\forall v \in \mathcal{V}$. To train an empirical graph model end-to-end, we used a graph-based cross-entropy loss function. 

\[\mathcal{L}_{G} = -\sum_{l \in \mathcal{Y}_{L}}y_{l}^{T} \rm ln(softmax(z^{(L)}_{v})) \]

\noindent where $\mathcal{Y}_{L}$ is the set of labeled nodes, and $y_{l}$ the ground truth represented by the one-hot encoding.

 \subsection{Statistical analysis}\label{subsec3}
 
Continuous variables were expressed as mean ± SD or median (range) for normally and non-normally distributed data, respectively. To assess normal distribution, the Kolmogorov-Smirnov test was performed. Continuous variables were compared using Student t-test or Wilcoxon rank sum test, and categorical variables were compared using Pearson's Chi-squared or Fisher's exact test as appropriate. A $p$-value $<$0.05 was considered statistically significant.

The performance of the transfer risk classification was analyzed by sensitivity (SEC), specificality (SPE), the receiver operating characteristic curve (ROC), and measured by the area under the ROC curve (AUC). Local Clustering Coefficient (LCC) was used to measure node connectivity in graph. All statistical analysis was performed using RStudio IDE tools (https://www.rstudio.com/) or Python SciPy package (www.scipy.org).

 \subsection{Experimental settings}

The entire study cohort was partitioned into a training set (60\%) and a testing set (40\%) using stratified random sampling. The choice of 60/40 partitioning, instead of the conventional 80/20 or 70/30, was based on the limited number of positive cases ($<$12\%) in our dataset. Baseline ML methods compared included Linear Discriminant Analysis (LDA), Support Vector Machine (SVM), Random Forest (RF), K-nearest neighbors (KNN) and Logistic regression (LR). The features of different days were concatenated for the ML based models. Since the original GraphSAGE model relied on data naturally containing linkages between objects, e.g., social networks and protein interactions, it was not directly comparable on our patient data. Therefore, we designed three versions of GraphSAGE-based GCN models for the evaluation as follows.

\begin{itemize}
	\item{KNN-GCN: It used K-neighbors method to find $K$ nearest neighbors for every given node and to construct a daily patient similarity graph. For a Day $T$ prediction, the final graph was obtained by aggregating graphs of Day 1, 2, ..., $T$, based on which a GCN was trained for classification.}
	\item{Diffusion-GCN: It used an additional diffusion process to aggregate daily patient similarity graphs, based on which a GCN was trained for classification.}
	\item {Diffusion-GCN (+age): It used Diffusion-GCN with age as an additional feature, included as node attribute for training.}
\end{itemize}

The number of clusters $K$ for the K-neighbors method used to build the adjacency graphs in our approach was experimentally set to $K$=200 for the training set and $K$=100 for the validation set, due to limited data size in the validation set. For the settings of GCNs, the mean aggregator was used as the aggregator, the number of hidden layers was set to 2, and each layer size was 62. Unlike traditional GCN methods, we subsampled 50 random nodes from the neighborhood for each hidden layer for training. To prevent overfitting, the dropout rate was set to 0.1. The network was trained using an Adam optimizer with a learning rate of 0.05 and an early stopping strategy. 

\subsection{Data availability}
The dataset generated during and/or analysed during the current study is not publicly available due to ethical and data safety reasons but is available from the corresponding author on reasonable request.

\begin{table}[ht]\large
\centering
\caption{\centering Summary statistics of the study population}

\begin{threeparttable}
\begin{tabular}{lcccc}
\hline
\textbf{Covariate}                                                        & \textbf{Non-severe (N=558)\tnote{1}} & \textbf{Severe (N=74)\tnote{1}} & \textbf{$p$-value\tnote{2}} \\ \hline
\textbf{Age}                                                              & 40 (19) [39, 42]                          & 51 (20) [46, 56]                     & \textless{}0.001  \\
\textbf{Gender}                                                           &                                           &                                      & 0.6               \\
\hspace{10pt}Female                                                                    & 273 (49) [45, 53]                   & 34 (46) [34, 58]             &                   \\
\hspace{10pt}Male                                                                      & 285 (51) [47, 55]                   & 40 (54) [42, 46]               &                   \\

\textbf{Vital signs}\tnote{3}                                                           &                                           &                                      &               \\

\hspace{10pt}Body Temperature ($^{\circ}$C)                              & 36 (0.63) [36, 36]                     & 37 (0.50) [36, 37]                & \textless{}0.001  \\
\hspace{10pt}SpO\textsubscript{2} (\%)                                          & 96 (1.11) [96, 96]                     & 95 (1.91) [95, 95]                & \textless{}0.001  \\
\hspace{10pt}Pulse rate (bpm)                                  & 88 (11) [87, 89]                          & 93 (11) [91, 96]                    & \textless{}0.001  \\
\hspace{10pt}Systolic blood pressure (mm Hg)                          & 85 (16) [84, 86]                          & 81 (10) [78, 83]                     & 0.1               \\
\hspace{10pt}Diastolic blood pressure (mm Hg)                         & 126 (14) [125, 127]                       & 127 (13) [124, 130]                  & 0.8               \\
\textbf{Comorbidities}                                                           &                                           &                                      &               \\

\hspace{10pt}Diabetes                                                         & 26 (4.7) [3.1, 6.8]                 & 11 (15.0) [8.0, 25.0]              & 0.002             \\
\hspace{10pt}Cardiovascular                                                   & 14 (2.5) [1.4, 4.3]                 & 5 (6.8) [2.5, 16.0]              & 0.06              \\
\hspace{10pt}Obesity                                                          & 19 (3.4) [2.1, 5.4]                 & 4 (5.4) [1.7, 14.0]              & 0.3               \\
\hspace{10pt}COPD\tnote{4}                  & 48 (8.6) [6.5, 11.0]                  & 8 (11.0) [5.1, 21.0]               & 0.5               \\

\textbf{Length of stay (day)}                                                      & 9 (3.0) [9.0, 9.5]                          & 3 (2.0) [2.2, 3.4]                     & \textless{}0.001  \\ \hline
\end{tabular}
\begin{tablenotes}
        \footnotesize               
       \item[1] n (\%); Mean (SD); 95\% Confidence Interval [lower limit, upper limit]
         \item[2] Pearson's Chi-squared test; Wilcoxon rank sum test; Fisher's exact test.
         \item[3] Computed based on the first three days of data.
         \item[4] COPD: Chronic Obstructive Pulmonary Disease.
      
\end{tablenotes}
\end{threeparttable}
\label{table1}
\end{table}

A total of 632 patients were included in this retrospective study, with a mean age of 41 $\pm$ 19 years old. Table \ref{table1} summarized the descriptive statistics of patient demographics, vital signs, and comorbidities. The mean age of the severe group was significantly higher than the non-severe group (51 vs. 40, $p<$0.001), and so was length of stay (3 vs. 9, $p<$0.001). Amongst five comorbidities, diabetes mellitus prevalence was significant between the two groups (11 vs. 26, $p<$0.002). Between the severe and non-severe groups, the differences were statistically significant for body temperature, SpO\textsubscript{2}, and pulse rate ($p<$0.001).

\begin{table}[b!]
\large
\centering
\caption{\centering  Model performances on Day 1, 2, and 3 predictions on the testing dataset}

\begin{threeparttable}
\begin{tabular}{lcccccccccccc}
 \multicolumn{1}{l}{} & \multicolumn{1}{l}{} & \multicolumn{1}{l}{} & \multicolumn{1}{l}{} & \multicolumn{1}{l}{} & \multicolumn{1}{l}{} & \multicolumn{1}{l}{} & \multicolumn{1}{l}{} & \multicolumn{1}{l}{} & \multicolumn{1}{l}{} \\ \hline
                           & \multicolumn{3}{c}{Day 1}                                           &                      & \multicolumn{3}{c}{Day 2}                                           &                      & \multicolumn{3}{c}{Day 3}                                           \\ \hline
                           & AUC                  & SEN          & SPE          &                      & AUC                  & SEN          & SPE          &                      & AUC                  & SEN          & SPE             \\ \cline{2-12} 
KNN-GCN              & 0.98                 & 1.00                    & 0.98                 &                      & 0.85                 & 0.67                 & 0.97                 &                      & 0.87                 & 0.63                 & 0.94                 \\
Diffusion-GCN        & 0.99                 & 0.75                 & 0.98                 &                      & 0.84                 & 0.50                  & 0.97                 &                      & 0.94                 & 0.75                 & 0.91                 \\
Diffusion-GCN ($+$age) & 0.99                 & 1.00                    & 0.98                 &                      & 0.83                 & 0.33                 & 0.97                 &                      & 0.89                 & 0.75                 & 0.93                 \\ \hline
LDA                        & 0.58                 & 0.00                    & 0.98                 &                      & 0.78                 & 0.33                 & 0.98                 &                      & 0.87                 & 0.63                 & 0.96                 \\
SVM                        & 0.72                 & 0.00                    & 0.99                 &                      & 0.79                 & 0.17                 & 0.99                 &                      & 0.88                 & 0.63                 & 0.96                 \\
RF             & 0.98                 & 1.00                    & 0.97                 &                      & 0.80                  & 0.50                  & 0.99                 &                      & 0.85                 & 0.50                  & 0.99                 \\
KNN                        & 0.99                 & 1.00                    & 0.92                 &                      & 0.64                 & 0.00                    & 1.00                    &                      & 0.88                 & 0.00                    & 1.00                    \\
LR        & 0.53                 & 0.00                    & 0.99                 &                      & 0.82                 & 0.33                 & 0.97                 &                      & 0.88                 & 0.50                  & 0.97                 \\ \hline
\end{tabular}
\begin{tablenotes}
    \small
    \item GCN: Graph Convolutional Networks; LDA: Linear Discriminant Analysis; SVM: Support Vector Machine; RF: Random Forest; KNN: K-nearest neighbors; LR: Logistic Regression; AUC: area under the ROC curve; SEN: Sensitivity; SPE: Specificity.
      
\end{tablenotes} 
\end{threeparttable}

\label{table2}
\end{table}

\subsection{Severity classification and model validation}

We evaluated the proposed GCN models for daily transfer risk classification against five popular ML methods. Table \ref{table2} compares model performances in terms of AUC, sensitivity, and specificity. Amongst ML methods, all except Random Forest suffered from zero sensitivity at some stage, which indicated that the models were totally biased towards the negative class. Random Forest was comparable to KNN-GCN and Diffusion-GCN(+age) on Day 1 prediction. However, starting from Day 2, GCN models overtook Random Forest, especially in AUC and sensitivity, achieving AUCs $\ge$0.83 for Day 1, 2, and 3 predictions. 

\begin{figure}[t!]
    \centering
    \includegraphics[width=1\linewidth]{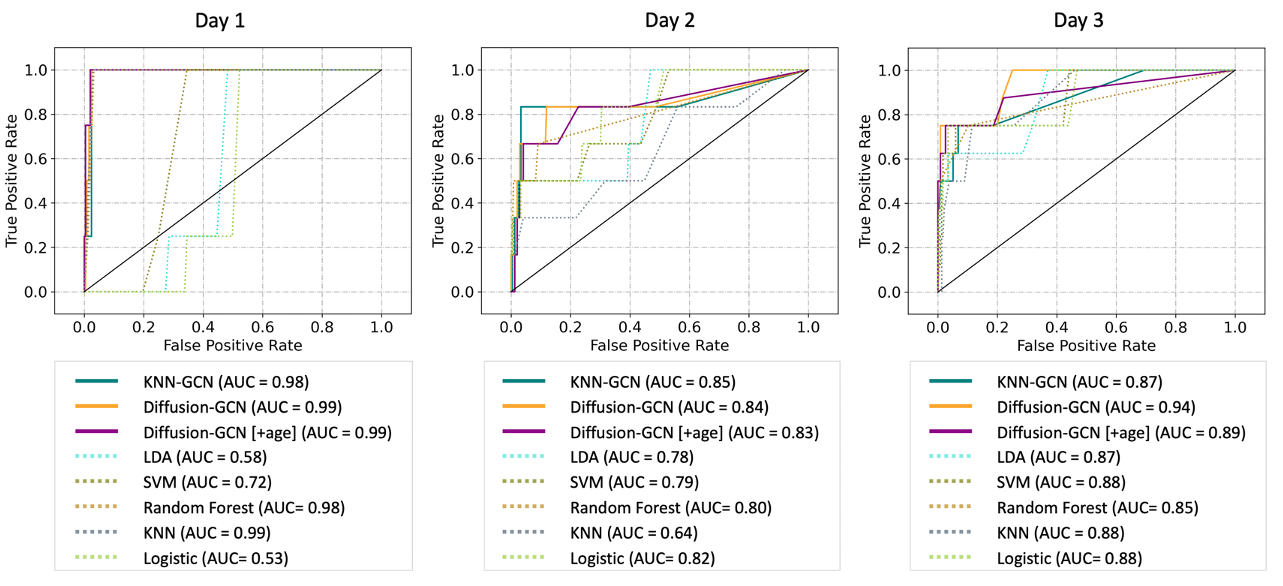}
    \caption{ROC curves of models predictions on Day 1, 2 and 3 on the testing dataset}
    \label{fig:Fig4}
 \end{figure}

Amongst GCN models, KNN-GCN and Diffusion-GCN were comparable. While KNN-GCN outperformed Diffusion-GCN on Day 1 and Day 2, especially in sensitivity ($\ge$0.67), Diffusion-GCN overtook KNN-GCN on Day 3 in both AUC (0.94) and sensitivity (0.75). Therefore, Diffusion-GCN demonstrated more stable performance for predictions based on more days of data. This can be explained by the ability of a diffusion process to aggregate progression over time. ROC curves in Fig. \ref{fig:Fig4} also demonstrated that the proposed GCN models outperformed baseline ML methods.

\subsection{Post-hoc graph analysis}

\begin{figure}[t!]
    \centering
    \includegraphics[width=0.99\linewidth]{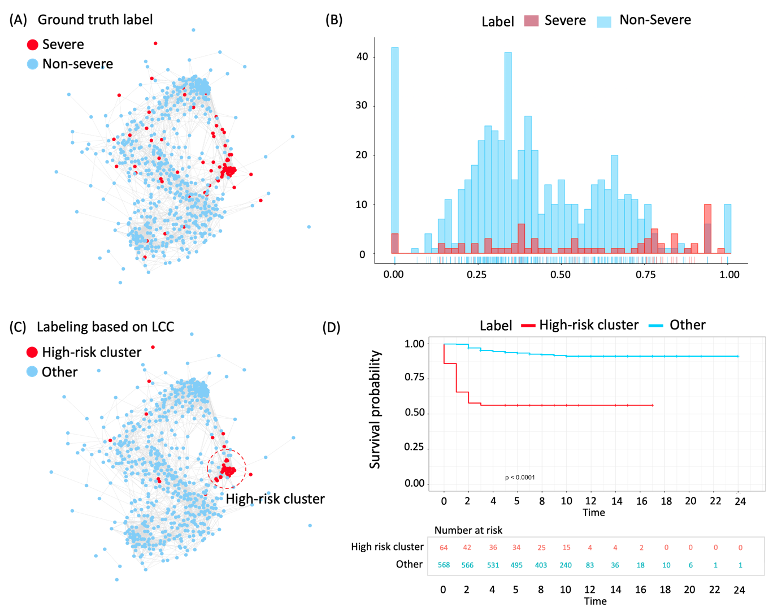}
    \caption{(A) Constructed diffusion-based patient similarity graph, with nodes colored according to the ground truth labels. (B) An LCC histogram calculated based on the graph. (C) Using the LCC cut-point to color the graph. (D) Time-to-hospital-transfer survival analysis on the discovered clusters.}
    \label{Fig5}
 \end{figure}

To further analyze the constructed diffusion-based graphs visually, we filtered out weights smaller than 0.014 for the graph on Day 3 and plotted a fully connected graph, with severe cases marked red and non-severe cases light blue based on the ground truth. As shown in Fig. \ref{Fig5} (A), a cluster of red points appears to be more concentrated.

 Based on the observation, we used Local Clustering Coefficient (LCC) \cite{LCC} to analyze the degree of aggregation between nodes on the graph to find concentrated clusters. The histogram of the resulting LCC coefficients was plotted in Fig. \ref{Fig5} (B), with the severe group marked in red and non-severe group in light blue. In general, the higher the LCC coefficient, the higher the degree of interconnection between the nodes. With a cutoff of LCC =0.75, we were able to identify a densely connected cluster of severe cases on the graph, as shown in the red circle in Fig. \ref{Fig5} (C). Comparing the summary statistics of identified high-risk cluster and the remaining cluster, Table \ref{table3} shows that the patients in the high-risk cluster were significantly older (54 vs. 41 years old, $p=$0.003), with significantly higher BT (36.57 vs. 36.21, $p<$0.001), lower SpO\textsubscript{2} (95.10 vs. 96.03, $p<$0.05), and shorter length of stay (1 vs. 9 days, $p<$0.001). Additionally, the LCC of the high-risk cluster was significantly higher (0.86 vs. 0.41, $p<$0.001).
\\

We also conducted survival analysis using the cox survival model for the clusters. Fig. \ref{Fig5} (D) shows the Kaplan-Meier curves for the clusters. The identified high-risk severe cluster (red) had the highest risk, with the time-to-hospital-transfer survival probability lower than 60\% in two days. For the remaining cluster (green), the survival probabilities showed a relatively stable step probability function decreasing over time.

\begin{table}[t!]
\caption{\centering  Summary statistics of LCC-discovered clusters}
\normalsize
\centering

\begin{threeparttable}
\begin{tabular}{lccc}
\hline
                    & High-risk (N$=$28)\tnote{1} & Other (N$=$604)\tnote{1} & $p$-value\tnote{2}            \\ \hline
\textbf{Age}                          & 54 (21)                  & 41 (19)                   & 0.003                \\
\textbf{Gender}                         &                          &                           &     0.113                 \\
\hspace{10pt}Female                       & 9 (32)                 & 298 (49)                &                      \\
\hspace{10pt}Male                         & 19 (68)                & 306 (51)                &                      \\
\textbf{Vital signs}                                                           &                                           &                                      &               \\

\hspace{10pt}Body Temperature (BT) ($^{\circ}$C)     & 36.57 (0.49)      & 36.21 (0.63)   & \textless{}0.001 \\
\hspace{10pt}SpO\textsubscript{2} (\%)                        & 95.10 (2.40)      & 96.03 (1.20)   & 0.049            \\
\hspace{10pt}Pulse rate (bpm)                 & 91 (8)            & 89 (11)        & 0.12             \\
\hspace{10pt}Systolic blood pressure (mm Hg)  & 83 (8)            & 84 (16)        & 0.4              \\
\hspace{10pt}Diastolic blood pressure (mm Hg) & 127 (11)          & 126 (14)       & 0.6              \\

\textbf{Local clustering coefficient} & 0.86 (0.07)              & 0.41 (0.22)               & \textless{}0.001     \\
\textbf{Length of stay (day)}                  & 1 (1)                    & 9 (3)                     & \textless{}0.001     \\ \hline
\end{tabular}
\begin{tablenotes}

        \footnotesize               
        \small \item[1] Mean (SD); n (\%).          
        \small \item[2] Wilcoxon rank sum test; Pearson's Chi-squared test.

\end{tablenotes}   
\end{threeparttable}
\label{table3}

\end{table}

\section{Discussion}
During the outbreak of a massive infectious disease, an accurate and easily-available risk stratification system was necessary to find potential critically ill patients and assign scarce medical resources to those in need \cite{Machine_Severity,Bolourani, Journal_Medical}. 
According to COVID-19 treatment guidelines from WHO \cite{Underlying_Medical}, remdesivir and dexamethasone were recommended to prevent disease progression for patients needing hospitalization and oxygen supplement. However, the timing of giving remdesivir is important \cite{Siemieniukm2980}. Therefore, a timely and accurate predictive system may help identify patients in need.

In this study, we developed and evaluated a graph-based deep learning approach to distinguishing severe COVID-19 patients who needed to be transferred to a hospital for further medical and respiratory support. Overall, the proposed GCN models outperformed all the ML models compared, demonstrating their superior abilities to differentiate between severe and non-severe cases using a graph representation (all AUCs $\ge$0.83). Although the proposed KNN-GCN and Diffusion-GCN were comparable, Diffusion-GCN demonstrated more stable performance for predictions based on more days of data.

Identifying high-risk COVID-19 patients was important to allocate limited medical resources during an outbreak. CURB-65 \cite{Lim377} and qSOFA \cite{Seymour} were commonly used predictive tools for estimating mortality for community-acquired pneumonia and suspected sepsis respectively. Haimovich et al developed COVID-19 severity index (CSI) using XGBoost algorithm and quick COVID-severity index (qCSI) using logistic regression, based on respiratory rate, pulse oximetry, and oxygen flow rate \cite{HAIMOVICH2020442}. The reported AUCs for CSI (76\%) and qCSI (81\%) outperformed classic CURB65 (50\%) and qSOFA (59\%) based on a dataset of 1792 patients who might develop respiratory failure in the emergent department.  However, unlike our progressive prediction approach, their model predicted at the 4 hour mark after admission and did not consider disease progression over time. 

The findings in our study echoed existing medical research. Our data showed that most severe patients became progressively ill within 3-4 days of their stay, which corresponded to the mean time from symptom onset to hospitalization reported in the study by Pellis et al (2.62 days in Singapore, 4.41 days in Hong Kong, and 5.14 days in the UK)\cite{short_doubling}. A large cross-sectional study by Kompaniyets et al based on more than 540,000 adults hospitalized with COVID-19 \cite{Underlying_Medical} found that the death risk ratio increased from 3.4 to 18.5 when the patients' age group changed from 40-49 to 50-64. Our study also discovered a high-risk cluster that had significantly older mean age (54.2 vs. 40.9, $p<$0.01). 
 
As the recent prevalent SARS-CoV-2 omicron strain caused less severe outcomes than previously dominant variants, public health policies for COVID-19 management started to encourage community-based health care rather than hospital-based treatment. However, unlike our work, most existing studies relied on the radiologic and laboratory data, which was not accessible in a community-based health care center, like the medicalized hotel in this study. 

Furthermore, our study demonstrated the feasibility of continuous remote patient monitoring echoing the studies of  \cite{Downey,Larimer}, as the data used in this study was collected via a virtual cloud-based ward care system, which was an extension from current hospital information system and could be easily implemented. The limitation of the current study lies in that the majority of the severe patients were transferred to a hospital within the first three days; therefore, there was not enough positive cases for training robust models for predictions over a longer period, which a larger dataset in the future may help.

\section{Conclusion}

Identifying potentially severe COVID-19 patients who needed to be transferred from a medicalized hotel to a hospital is critical for timely medical and respiratory support. In this study, we developed and evaluated a graph-based progressive hospital transfer risk prediction approach based on daily progression of vital signs and accumulated transition probability. Although this study was based on patient data when the alpha-variant was the dominant COVID-19 strain, our approach demonstrated superior performance and potential applicability to future prediction tasks with limited longitudinal measurements at hand, as in the case of a medicalized hotel.  
\newline

\bibliographystyle{unsrtnat}
\bibliography{references}  

\begin{thebibliography}{49}
\providecommand{\natexlab}[1]{#1}
\providecommand{\url}[1]{\texttt{#1}}
\expandafter\ifx\csname urlstyle\endcsname\relax
  \providecommand{\doi}[1]{doi: #1}\else
  \providecommand{\doi}{doi: \begingroup \urlstyle{rm}\Url}\fi

\bibitem[Chen et~al.(2020)Chen, Zhang, Yang, Wang, Zhai, Bärnighausen, and
  Wang]{Fangcang}
Simiao Chen, Zongjiu Zhang, Juntao Yang, Jian Wang, Xiaohui Zhai, Till
  Bärnighausen, and Chen Wang.
\newblock {Fangcang shelter hospitals: a novel concept for responding to public
  health emergencies}.
\newblock \emph{Lancet (London, England)}, 395\penalty0 (10232):\penalty0
  1305--1314, 2020.
\newblock ISSN 0140-6736.
\newblock \doi{https://doi.org/10.1016/s0140-6736(20)30744-3}.

\bibitem[Brady et~al.(2021)Brady, Milzman, Walton, Sommer, Neustadtl, and
  Napoli]{10.1093/milmed/usab003}
Kevin Brady, Dave Milzman, Edward Walton, Darren Sommer, Alan Neustadtl, and
  Anthony Napoli.
\newblock {Uniformed Services and the Field Hospital Experience During
  Coronovirus Disease 2019 (SARS-CoV-2) Pandemic: Open to Closure in 30 Days
  With 1,100 Patients: The Javits New York Medical Station}.
\newblock \emph{Military Medicine}, pages usab003--, 2021.
\newblock ISSN 1930-613X.
\newblock \doi{https://doi.org/10.1093/milmed/usab003}.

\bibitem[Spagnolello et~al.(2020)Spagnolello, Rota, Valoti, Cozzini, Parrino,
  Portella, and Langer]{Bergamo}
Ornella Spagnolello, Silvia Rota, Oliviero~Francesco Valoti, Claudio Cozzini,
  Pietro Parrino, Gina Portella, and Martin Langer.
\newblock {Bergamo Field Hospital Confronting COVID-19: Operating
  Instructions}.
\newblock \emph{Disaster Medicine and Public Health Preparedness}, pages 1--3,
  2020.
\newblock ISSN 1935-7893.
\newblock \doi{https://doi.org/10.1017/dmp.2020.447}.

\bibitem[Bulajic et~al.(2021)Bulajic, Ekambaram, Saunders, Naidoo, Wallis,
  Amien, Ras, Pressentin, Tadzimirwa, Hussey, Reid, and Hodkinson]{Bulajic}
Bojana Bulajic, Kamlin Ekambaram, Colleen Saunders, Vanessa Naidoo, Lee Wallis,
  Nabeela Amien, Tasleem Ras, Klaus~von Pressentin, Gamuchirai Tadzimirwa,
  Nadia Hussey, Steve Reid, and Peter Hodkinson.
\newblock {A COVID-19 field hospital in a conference centre – The Cape Town,
  South Africa experience}.
\newblock \emph{African Journal of Primary Health Care \& Family Medicine},
  2021.
\newblock ISSN 2071-2928.
\newblock \doi{https://doi.org/10.4102/phcfm.v13i1.3140}.

\bibitem[Baughman et~al.(2020)Baughman, Hirschberg, Lucas, Suarez, Stockmann,
  Johnson, Hutter, Murphy, Marsh, Thompson, Boland, Erickson, and
  Palamara]{Pandemic_Care}
Amy~W. Baughman, Ronald~E. Hirschberg, Larissa~J. Lucas, Elliot~D. Suarez,
  Deanna Stockmann, Stacy~Hutton Johnson, Matthew~M. Hutter, Deborah~J. Murphy,
  Regan~H. Marsh, Ryan~W. Thompson, Giles~W. Boland, Jeanette~Ives Erickson,
  and Kerri Palamara.
\newblock {Pandemic Care Through Collaboration: Lessons From a COVID-19 Field
  Hospital}.
\newblock \emph{Journal of the American Medical Directors Association},
  21\penalty0 (11):\penalty0 1563--1567, 2020.
\newblock ISSN 1525-8610.
\newblock \doi{https://doi.org/10.1016/j.jamda.2020.09.003}.

\bibitem[Kompaniyets et~al.(2021)Kompaniyets, Pennington, Goodman, and
  Rosenblum]{Underlying_Medical}
Lyudmyla Kompaniyets, Audrey~F Pennington, Alyson~B Goodman, and Hannah G et~al
  Rosenblum.
\newblock {Underlying Medical Conditions and Severe Illness Among 540,667
  Adults Hospitalized With COVID-19, March 2020-March 2021}.
\newblock \emph{Preventing Chronic Disease}, 18:\penalty0 E66, 2021.
\newblock ISSN 1545-1151.
\newblock \doi{https://doi.org/10.5888/pcd18.210123}.

\bibitem[James et~al.(2013)James, Witten, Hastie, and Tibshirani]{ISLR}
Gareth James, Daniela Witten, Trevor Hastie, and Robert Tibshirani.
\newblock \emph{An Introduction to Statistical Learning}.
\newblock Springer, 01 2013.
\newblock ISBN 1461471389.

\bibitem[Haimovich et~al.(2020{\natexlab{a}})Haimovich, Ravindra, Stoytchev,
  Young, Wilson, Dijk, Schulz, and Taylor]{Annals_Haimovich}
Adrian~D. Haimovich, Neal~G. Ravindra, Stoytcho Stoytchev, H.~Patrick Young,
  Francis~P. Wilson, David~van Dijk, Wade~L. Schulz, and R.~Andrew Taylor.
\newblock {Development and Validation of the Quick COVID-19 Severity Index: A
  Prognostic Tool for Early Clinical Decompensation}.
\newblock \emph{Annals of Emergency Medicine}, 76\penalty0 (4):\penalty0
  442--453, 2020{\natexlab{a}}.
\newblock ISSN 0196-0644.
\newblock \doi{https://doi.org/10.1016/j.annemergmed.2020.07.022}.

\bibitem[Wu et~al.(2020)Wu, Yang, Xie, Woodruff, and Rao]{Wu2001104}
Guangyao Wu, Pei Yang, Yuanliang Xie, Henry~C. Woodruff, and Xiangang et~al
  Rao.
\newblock Development of a clinical decision support system for severity risk
  prediction and triage of covid-19 patients at hospital admission: an
  international multicentre study.
\newblock \emph{European Respiratory Journal}, 56\penalty0 (2), 2020.
\newblock ISSN 0903-1936.
\newblock \doi{https://doi.org/10.1183/13993003.01104-2020}.

\bibitem[Patel et~al.(2020{\natexlab{a}})Patel, Kher, Desai, Lei, Cen, Nanda,
  Gholamrezanezhad, Duddalwar, Varghese, and Oberai]{MLCOVID}
Dhruv Patel, Vikram Kher, Bhushan Desai, Xiaomeng Lei, Steven Cen, Neha Nanda,
  Ali Gholamrezanezhad, Vinay Duddalwar, Bino Varghese, and Assad Oberai.
\newblock Machine learning based predictors for covid-19 disease severity.
\newblock 11 2020{\natexlab{a}}.
\newblock \doi{10.21203/rs.3.rs-108301/v1}.

\bibitem[Schwab et~al.(2020)Schwab, Sch{\"{u}}tte, Dietz, and
  Bauer]{predCOVID-19}
Patrick Schwab, August~DuMont Sch{\"{u}}tte, Benedikt Dietz, and Stefan Bauer.
\newblock predcovid-19: {A} systematic study of clinical predictive models for
  coronavirus disease 2019.
\newblock \emph{CoRR}, abs/2005.08302, 2020.

\bibitem[Gao et~al.(2020)Gao, Cai, Fang, Li, Wang, and Chen]{mortality_risk}
Yue Gao, Guangyao Cai, Wei Fang, Hua-Yi Li, Si-Yuan Wang, and Lingxi et~al
  Chen.
\newblock Machine learning based early warning system enables accurate
  mortality risk prediction for covid-19.
\newblock \emph{Nature Communications}, 11:\penalty0 1--10, 10 2020.
\newblock \doi{https://doi.org/10.1038/s41467-020-18684-2}.

\bibitem[Hu et~al.(2020)Hu, Liu, Jiang, Shi, Zhang, and et~al]{Epidemiology}
Chuanyu Hu, Zhenqiu Liu, Yanfeng Jiang, Oumin Shi, Xin Zhang, and et~al.
\newblock {Early prediction of mortality risk among patients with severe
  COVID-19, using machine learning}.
\newblock \emph{International Journal of Epidemiology}, 49\penalty0
  (6):\penalty0 1918--1929, 09 2020.
\newblock ISSN 0300-5771.
\newblock \doi{https://doi.org/10.1093/ije/dyaa171}.

\bibitem[Nemati et~al.(2020)Nemati, Ansary, and Nemati]{NEMATI2020100074}
Mohammadreza Nemati, Jamal Ansary, and Nazafarin Nemati.
\newblock Machine-learning approaches in covid-19 survival analysis and
  discharge-time likelihood prediction using clinical data.
\newblock \emph{Patterns}, 1\penalty0 (5):\penalty0 100074, 2020.
\newblock ISSN 2666-3899.
\newblock \doi{https://doi.org/10.1016/j.patter.2020.100074}.

\bibitem[{Di Castelnuovo} et~al.(2020){Di Castelnuovo}, Bonaccio, Costanzo, and
  et~al]{DICASTELNUOVO20201899}
Augusto {Di Castelnuovo}, Marialaura Bonaccio, Simona Costanzo, and
  Alessandro~Gialluisi et~al.
\newblock Common cardiovascular risk factors and in-hospital mortality in 3,894
  patients with covid-19: survival analysis and machine learning-based findings
  from the multicentre italian corist study.
\newblock \emph{Nutrition, Metabolism and Cardiovascular Diseases}, 30\penalty0
  (11):\penalty0 1899--1913, 2020.
\newblock ISSN 0939-4753.
\newblock \doi{https://doi.org/10.1016/j.numecd.2020.07.031}.

\bibitem[Goodfellow et~al.(2016)Goodfellow, Bengio, and Courville]{Goodfellow}
Ian Goodfellow, Yoshua Bengio, and Aaron Courville.
\newblock \emph{Deep Learning}.
\newblock MIT Press, 2016.

\bibitem[Mohamadou et~al.(2020)Mohamadou, Halidou, and Kapen]{Mohamadou2020}
Youssoufa Mohamadou, Aminou Halidou, and Pascalin~Tiam Kapen.
\newblock {A review of mathematical modeling, artificial intelligence and
  datasets used in the study, prediction and management of COVID-19}.
\newblock \emph{Applied Intelligence}, 50:\penalty0 3913--3925, 2020.

\bibitem[Jangam et~al.(2022)Jangam, Barreto, and Annavarapu]{Jangam2022}
Ebenezer Jangam, Aaron Antonio~Dias Barreto, and Chandra Sekhara~Rao
  Annavarapu.
\newblock {Automatic detection of COVID-19 from chest CT scan and chest X-Rays
  images using deep learning, transfer learning and stacking}.
\newblock \emph{Applied Intelligence}, 52\penalty0 (2):\penalty0 2243--2259,
  2022.
\newblock ISSN 15737497.
\newblock \doi{https://doi.org/10.1007/s10489-021-02393-4}.

\bibitem[Qi et~al.(2021)Qi, Xu, Li, Tian, Xia, Ren, Yang, Wang, and
  Yu]{QI2021106406}
Shouliang Qi, Caiwen Xu, Chen Li, Bin Tian, Shuyue Xia, Jigang Ren, Liming
  Yang, Hanlin Wang, and Hui Yu.
\newblock Dr-mil: deep represented multiple instance learning distinguishes
  covid-19 from community-acquired pneumonia in ct images.
\newblock \emph{Computer Methods and Programs in Biomedicine}, 211:\penalty0
  106406, 2021.
\newblock ISSN 0169-2607.
\newblock \doi{https://doi.org/10.1016/j.cmpb.2021.106406}.

\bibitem[Lassau et~al.(2021)Lassau, Ammari, Chouzenoux, Gortais, Herent,
  Devilder, Soliman, Meyrignac, Talabard, Lamarque, Dubois, Loiseau,
  Trichelair, Bendjebbar, Garcia, Balleyguier, Merad, Stoclin, Jegou, and
  Blum]{Integrating}
Nathalie Lassau, S.~Ammari, Emilie Chouzenoux, Hugo Gortais, Paul Herent,
  Matthieu Devilder, Samer Soliman, Olivier Meyrignac, Marie-Pauline Talabard,
  Jean-Philippe Lamarque, Remy Dubois, Nicolas Loiseau, Paul Trichelair,
  Etienne Bendjebbar, Gabriel Garcia, Corinne Balleyguier, Mansouria Merad,
  Annabelle Stoclin, Simon Jegou, and Michael Blum.
\newblock Integrating deep learning ct-scan model, biological and clinical
  variables to predict severity of covid-19 patients.
\newblock \emph{Nature Communications}, 12:\penalty0 634, 01 2021.
\newblock \doi{https://doi.org/10.1038/s41467-020-20657-4}.

\bibitem[Khan et~al.(2020)Khan, Shah, and Bhat]{KHAN2020105581}
Asif~Iqbal Khan, Junaid~Latief Shah, and Mohammad~Mudasir Bhat.
\newblock Coronet: A deep neural network for detection and diagnosis of
  covid-19 from chest x-ray images.
\newblock \emph{Computer Methods and Programs in Biomedicine}, 196:\penalty0
  105581, 2020.
\newblock ISSN 0169-2607.
\newblock \doi{https://doi.org/10.1016/j.cmpb.2020.105581}.

\bibitem[P and Annavarapu(2021)]{P2021}
Samson Anosh~Babu P and Chandra Sekhara~Rao Annavarapu.
\newblock {Deep learning-based improved snapshot ensemble technique for
  COVID-19 chest X-ray classification}.
\newblock \emph{Applied Intelligence}, 51\penalty0 (5):\penalty0 3104--3120,
  2021.
\newblock ISSN 15737497.
\newblock \doi{https://doi.org/10.1007/s10489-021-02199-4}.

\bibitem[Li et~al.(2020)Li, Ge, Zhu, Li, Graham, Singer, Richman, and
  Duong]{Incremental}
Xiaoran Li, Peilin Ge, Jocelyn Zhu, Haifang Li, James Graham, Adam Singer, Paul
  Richman, and Tim Duong.
\newblock Deep learning prediction of likelihood of icu admission and mortality
  in covid-19 patients using clinical variables.
\newblock \emph{PeerJ}, 8:\penalty0 e10337, 11 2020.
\newblock \doi{https://doi.org/10.7717/peerj.10337}.

\bibitem[Li and Patra(2010)]{Genome-wide}
Yongjin Li and Jagdish Patra.
\newblock Genome-wide inferring gene-phenotype relationship by walking on the
  heterogeneous network.
\newblock \emph{Bioinformatics (Oxford, England)}, 26:\penalty0 1219--24, 03
  2010.
\newblock \doi{https://doi.org/10.1093/bioinformatics/btq108}.

\bibitem[Wang et~al.(2014)Wang, Mezlini, Demir, Fiume, Tu, Brudno, Haibe-Kains,
  and Goldenberg]{Wang2014SimilarityNF}
Bo~Wang, Aziz~M. Mezlini, Feyyaz Demir, Marc Fiume, Zhuowen Tu, Michael Brudno,
  Benjamin Haibe-Kains, and Anna Goldenberg.
\newblock Similarity network fusion for aggregating data types on a genomic
  scale.
\newblock \emph{Nature Methods}, 11:\penalty0 333--337, 2014.

\bibitem[Kipf and Welling(2016)]{KipfW16}
Thomas~N. Kipf and Max Welling.
\newblock Semi-supervised classification with graph convolutional networks.
\newblock \emph{CoRR}, abs/1609.02907, 2016.

\bibitem[Keicher et~al.(2021)Keicher, Burwinkel, Bani-Harouni, Paschali,
  Czempiel, Burian, Makowski, Braren, Navab, and Wendler]{keicher2021ugat}
Matthias Keicher, Hendrik Burwinkel, David Bani-Harouni, Magdalini Paschali,
  Tobias Czempiel, Egon Burian, Marcus~R. Makowski, Rickmer Braren, Nassir
  Navab, and Thomas Wendler.
\newblock U-gat: Multimodal graph attention network for covid-19 outcome
  prediction.
\newblock 2021.

\bibitem[Yu et~al.(2021)Yu, Lu, Guo, Wang, and Zhang]{YU2021592}
Xiang Yu, Siyuan Lu, Lili Guo, Shui-Hua Wang, and Yu-Dong Zhang.
\newblock Resgnet-c: A graph convolutional neural network for detection of
  covid-19.
\newblock \emph{Neurocomputing}, 452:\penalty0 592--605, 2021.
\newblock ISSN 0925-2312.
\newblock \doi{https://doi.org/10.1016/j.neucom.2020.07.144}.

\bibitem[Ferrari et~al.(2022)Ferrari, Gargani, Barbieri, Ghiadoni, Faita, and
  Bacciu]{Ferrari}
Elisa Ferrari, Luna Gargani, Greta Barbieri, Lorenzo Ghiadoni, Francesco Faita,
  and Davide Bacciu.
\newblock A causal learning framework for the analysis and interpretation of
  covid-19 clinical data.
\newblock \emph{PLOS ONE}, 17\penalty0 (5):\penalty0 1--21, 05 2022.
\newblock \doi{https://doi.org/10.1371/journal.pone.0268327}.

\bibitem[Krieg et~al.(2021)Krieg, Avendano, Grantham-Brown, Asbun, Schnur,
  Miranda, and Chawla]{KRI}
Steven~J. Krieg, Carolina Avendano, Evan Grantham-Brown, Aaron~Lilienfeld
  Asbun, Jennifer~J. Schnur, Marie~Lynn Miranda, and Nitesh~V. Chawla.
\newblock Data-driven testing program improves detection of covid-19 cases and
  reduces community transmission.
\newblock \emph{medRxiv}, 2021.
\newblock \doi{https://doi.org/10.1101/2021.07.31.21261423}.

\bibitem[Ko et~al.(2020)Ko, Chung, Kang, Kim, Shin, Kang, Lee, Kim, Kim, Jung,
  and Lee]{pneumonia_diagnosis}
Hoon Ko, Heewon Chung, Wu~Seong Kang, Kyung~Won Kim, Youngbin Shin, Seung~Ji
  Kang, Jae~Hoon Lee, Young~Jun Kim, Nan~Yeol Kim, Hyunseok Jung, and Jinseok
  Lee.
\newblock Covid-19 pneumonia diagnosis using a simple 2d deep learning
  framework with a single chest ct image: Model development and validation.
\newblock \emph{J Med Internet Res}, 22\penalty0 (6):\penalty0 e19569, Jun
  2020.
\newblock ISSN 1438-8871.
\newblock \doi{https://doi.org/10.2196/19569}.

\bibitem[Zhao et~al.(2021)Zhao, Jiang, and Qiu]{covid_detection}
Wentao Zhao, Wei Jiang, and Xinguo Qiu.
\newblock Deep learning for covid-19 detection based on ct images.
\newblock \emph{Scientific Reports}, 11, 07 2021.
\newblock \doi{https://doi.org/10.1038/s41598-021-93832-2}.

\bibitem[Huang et~al.(2021)Huang, Lin, Zhang, Xu, Zheng, Mao, Qian, Peng, Zhou,
  Chen, and Tong]{Boundary}
Huimin Huang, Lanfen Lin, Yue Zhang, Yingying Xu, Jing Zheng, XiongWei Mao,
  Xiaohan Qian, Zhiyi Peng, Jianying Zhou, Yen-Wei Chen, and Ruofeng Tong.
\newblock Graph-bas3net: Boundary-aware semi-supervised segmentation network
  with bilateral graph convolution.
\newblock In \emph{Proceedings of the IEEE/CVF International Conference on
  Computer Vision (ICCV)}, pages 7386--7395, October 2021.

\bibitem[Lim et~al.(2021)Lim, Chen, Chiu, and Hung]{LIM2021}
Kai~Xuan Lim, Yi-Tui Chen, Kuan-Ming Chiu, and Fang~Ming Hung.
\newblock Rush hour:transform a modern hotel into cloud-based virtual ward care
  center within 80 hours under covid-19 pandemic.far eastern memorialhospital's
  experience.
\newblock \emph{Journal of the Formosan Medical Association}, 2021.
\newblock ISSN 0929-6646.
\newblock \doi{https://doi.org/https://doi.org/10.1016/j.jfma.2021.10.023}.

\bibitem[van Buuren and Groothuis-Oudshoorn(2011)]{MICE}
Stef van Buuren and Karin Groothuis-Oudshoorn.
\newblock mice: Multivariate imputation by chained equations in r.
\newblock \emph{Journal of Statistical Software}, 45\penalty0 (3):\penalty0
  1–67, 2011.
\newblock \doi{https://doi.org/10.18637/jss.v045.i03}.

\bibitem[Wu et~al.(2021)Wu, Shen, Xu, and Shao]{dynamic}
Jiachao Wu, Jiang Shen, Man Xu, and Minglai Shao.
\newblock A novel combined dynamic ensemble selection model for imbalanced data
  to detect covid-19 from complete blood count.
\newblock \emph{Computer Methods and Programs in Biomedicine}, 211:\penalty0
  106444, 2021.
\newblock ISSN 0169-2607.
\newblock \doi{https://doi.org/10.1016/j.cmpb.2021.106444}.

\bibitem[Li et~al.(2018)Li, Yu, Shahabi, and Liu]{traffic}
Yaguang Li, Rose Yu, Cyrus Shahabi, and Yan Liu.
\newblock Diffusion convolutional recurrent neural network: Data-driven traffic
  forecasting.
\newblock In \emph{International Conference on Learning Representations (ICLR
  '18)}, 2018.

\bibitem[Hamilton et~al.(2017)Hamilton, Ying, and Leskovec]{NIPS2017_5dd9db5e}
Will Hamilton, Zhitao Ying, and Jure Leskovec.
\newblock Inductive representation learning on large graphs.
\newblock In I.~Guyon, U.~Von Luxburg, S.~Bengio, H.~Wallach, R.~Fergus,
  S.~Vishwanathan, and R.~Garnett, editors, \emph{Advances in Neural
  Information Processing Systems}, volume~30. Curran Associates, Inc., 2017.

\bibitem[Ferraz~de Arruda et~al.(2018)Ferraz~de Arruda, Rodrigues, and
  Moreno]{LCC}
Guilherme Ferraz~de Arruda, Francisco Rodrigues, and Yamir Moreno.
\newblock Fundamentals of spreading processes in single and multilayer complex
  networks.
\newblock \emph{Physics Reports}, 756, 04 2018.
\newblock \doi{https://doi.org/10.1016/j.physrep.2018.06.007}.

\bibitem[Patel et~al.(2020{\natexlab{b}})Patel, Kher, Desai, Lei, Cen, Nanda,
  Gholamrezanezhad, Duddalwar, Varghese, and Oberai]{Machine_Severity}
Dhruv Patel, Vikram Kher, Bhushan Desai, Xiaomeng Lei, Steven Cen, Neha Nanda,
  Ali Gholamrezanezhad, Vinay Duddalwar, Bino Varghese, and Assad Oberai.
\newblock Machine learning based predictors for covid-19 disease severity.
\newblock \emph{Scientific Reports}, 11 2020{\natexlab{b}}.
\newblock \doi{https://doi.org/10.21203/rs.3.rs-108301/v1}.

\bibitem[Bolourani et~al.(2020)Bolourani, Brenner, Wang, Mcginn, Hirsch,
  Barnaby, and Zanos]{Bolourani}
Siavash Bolourani, Max Brenner, Ping Wang, Thomas Mcginn, Jamie Hirsch, Douglas
  Barnaby, and Theodoros Zanos.
\newblock Development and validation of a machine learning prediction model of
  respiratory failure within 48 hours of patient admission for covid-19
  (preprint).
\newblock \emph{Journal of Medical Internet Research}, 23, 09 2020.
\newblock \doi{https://doi.org/10.2196/24246}.

\bibitem[Kim et~al.(2020)Kim, Han, Kim, Kim, Ha, Seog, Lee, Lim, Hong, Park,
  and Heo]{Journal_Medical}
Hyung-Jun Kim, Deokjae Han, Jeong-Han Kim, Daehyun Kim, Beomman Ha, Woong Seog,
  Yeon-Kyeng Lee, Dosang Lim, Sung~Ok Hong, Mi-Jin Park, and JoonNyung Heo.
\newblock {An Easy-to-Use Machine Learning Model to Predict the Prognosis of
  Patients With COVID-19: Retrospective Cohort Study}.
\newblock \emph{Journal of Medical Internet Research}, 22\penalty0
  (11):\penalty0 e24225, 2020.
\newblock \doi{https://doi.org/10.2196/24225}.

\bibitem[Siemieniuk et~al.(2020)Siemieniuk, Bartoszko, Ge, Zeraatkar, and
  Izcovich]{Siemieniukm2980}
Reed~AC Siemieniuk, Jessica~J Bartoszko, Long Ge, Dena Zeraatkar, and Ariel
  et~al Izcovich.
\newblock Drug treatments for covid-19: living systematic review and network
  meta-analysis.
\newblock \emph{BMJ}, 370, 2020.
\newblock \doi{https://doi.org/10.1136/bmj.m2980}.

\bibitem[Lim et~al.(2003)Lim, van~der Eerden, Laing, Boersma, Karalus, Town,
  Lewis, and Macfarlane]{Lim377}
W~S Lim, M~M van~der Eerden, R~Laing, W~G Boersma, N~Karalus, G~I Town, S~A
  Lewis, and J~T Macfarlane.
\newblock Defining community acquired pneumonia severity on presentation to
  hospital: an international derivation and validation study.
\newblock \emph{Thorax}, 58\penalty0 (5):\penalty0 377--382, 2003.
\newblock ISSN 0040-6376.
\newblock \doi{https://doi.org/10.1136/thorax.58.5.377}.

\bibitem[Seymour et~al.(2016)Seymour, Liu, Iwashyna, Brunkhorst, Rea, Scherag,
  Rubenfeld, Kahn, Shankar-Hari, Singer, Deutschman, Escobar, and
  Angus]{Seymour}
Christopher~W. Seymour, Vincent~X. Liu, Theodore~J. Iwashyna, Frank~M.
  Brunkhorst, Thomas~D. Rea, André Scherag, Gordon Rubenfeld, Jeremy~M. Kahn,
  Manu Shankar-Hari, Mervyn Singer, Clifford~S. Deutschman, Gabriel~J. Escobar,
  and Derek~C. Angus.
\newblock {Assessment of Clinical Criteria for Sepsis: For the Third
  International Consensus Definitions for Sepsis and Septic Shock (Sepsis-3)}.
\newblock \emph{JAMA}, 315\penalty0 (8):\penalty0 762--774, 02 2016.
\newblock ISSN 0098-7484.

\bibitem[Haimovich et~al.(2020{\natexlab{b}})Haimovich, Ravindra, Stoytchev,
  Young, Wilson, {van Dijk}, Schulz, and Taylor]{HAIMOVICH2020442}
Adrian~D. Haimovich, Neal~G. Ravindra, Stoytcho Stoytchev, H.~Patrick Young,
  Francis~P. Wilson, David {van Dijk}, Wade~L. Schulz, and R.~Andrew Taylor.
\newblock Development and validation of the quick covid-19 severity index: A
  prognostic tool for early clinical decompensation.
\newblock \emph{Annals of Emergency Medicine}, 76\penalty0 (4):\penalty0
  442--453, 2020{\natexlab{b}}.
\newblock ISSN 0196-0644.
\newblock \doi{https://doi.org/10.1016/j.annemergmed.2020.07.022}.

\bibitem[Pellis et~al.(2021)Pellis, Scarabel, Stage, Overton, Chappell, Fearon,
  Bennett, Lythgoe, House, Hall, and null]{short_doubling}
Lorenzo Pellis, Francesca Scarabel, Helena~B. Stage, Christopher~E. Overton,
  Lauren H.~K. Chappell, Elizabeth Fearon, Emma Bennett, Katrina~A. Lythgoe,
  Thomas~A. House, Ian Hall, and null null.
\newblock Challenges in control of covid-19: short doubling time and long delay
  to effect of interventions.
\newblock \emph{Philosophical Transactions of the Royal Society B: Biological
  Sciences}, 376\penalty0 (1829):\penalty0 20200264, 2021.
\newblock \doi{https://doi.org/10.1098/rstb.2020.0264}.

\bibitem[Downey et~al.(2018)Downey, Randell, Brown, and Jayne]{Downey}
Candice Downey, Rebecca Randell, Julia Brown, and David~G Jayne.
\newblock Continuous versus intermittent vital signs monitoring using a
  wearable, wireless patch in patients admitted to surgical wards: Pilot
  cluster randomized controlled trial.
\newblock \emph{J Med Internet Res}, 20\penalty0 (12):\penalty0 e10802, Dec
  2018.

\bibitem[Larimer et~al.(2021)Larimer, Wegerich, Splan, Chestek, Prendergast,
  and Vanden~Hoek]{Larimer}
Karen Larimer, Stephan Wegerich, Joel Splan, David Chestek, Heather
  Prendergast, and Terry Vanden~Hoek.
\newblock Personalized analytics and a wearable biosensor platform for early
  detection of covid-19 decompensation (decode): Protocol for the development
  of the covid-19 decompensation index.
\newblock \emph{JMIR Res Protoc}, 10\penalty0 (5):\penalty0 e27271, May 2021.

\end{thebibliography}






\end{document}